\documentclass[sigconf,natbib=false, anonymous=false, balance=false]{acmart}

\AtBeginDocument{%
  }

\usepackage[LGR,T1]{fontenc}
  \newcommand{\textgreek}[1]{\begingroup\fontencoding{LGR}\selectfont#1\endgroup}
\usepackage[utf8]{inputenc}
\usepackage{subcaption}

\setcopyright{acmlicensed}
\copyrightyear{2024}
\acmYear{2024}
\acmDOI{XXXXXXX.XXXXXXX}

\usepackage[numbers]{natbib}

\ccsdesc[500]{General and reference~General conference proceedings}

\begin{document}

\acmConference[SETN '24]{The 13th EETN Conference on Artificial Intelligence}{September 11--13,
  2024}{Piraeus, Greece}

\title{Greek2MathTex: A Greek Speech-to-Text Framework for LaTeX Equations Generation}

\author{Evangelia Gkritzali}
\email{evagkri@gmail.com}
\affiliation{%
  \institution{NCSR Demokritos}
  \city{Athens}
  \state{Attiki}
  \country{Greece}
}

\author{Panagiotis Kaliosis}
\email{pkaliosis@iit.demokritos.gr}
\affiliation{%
  \institution{NCSR Demokritos}
  \city{Athens}
  \state{Attiki}
  \country{Greece}
}

\author{Sofia Galanaki}
\email{sgalan@iit.demokritos.gr}
\affiliation{%
  \institution{NCSR Demokritos}
  \city{Athens}
  \state{Attiki}
  \country{Greece}
}

\author{Elisavet Palogiannidi}
\email{epalogiannidi@iit.demokritos.gr}
\affiliation{%
  \institution{NCSR Demokritos}
  \city{Athens}
  \state{Attiki}
  \country{Greece}
}

\author{Theodoros Giannakopoulos}
\email{tyianak@iit.demokritos.gr}
\affiliation{%
  \institution{NCSR Demokritos}
  \city{Athens}
  \state{Attiki}
  \country{Greece}
}

\renewcommand{\shortauthors}{Gkritzali et al.}

\begin{abstract}

In the vast majority of the academic and scientific domains, \LaTeX\ has established itself as the de facto standard for typesetting complex mathematical equations and formulae. However, \LaTeX\'s complex syntax and code-like appearance present accessibility barriers for individuals with disabilities, as well as those unfamiliar with coding conventions. In this paper, we present a novel solution to this challenge through the development of a novel speech-to-\LaTeX\ equations system specifically designed for the Greek language. We propose an end-to-end system that harnesses the power of Automatic Speech Recognition (ASR) and Natural Language Processing (NLP) techniques to enable users to verbally dictate mathematical expressions and equations in natural language, which are subsequently converted into \LaTeX\ format. We present the architecture and design principles of our system, highlighting key components such as the ASR engine, the LLM-based prompt-driven equations generation mechanism, as well as the application of a custom evaluation metric employed throughout the development process. We have made our system open source and available at \href{https://github.com/magcil/greek-speech-to-math}{\textcolor{blue}{https://github.com/magcil/greek-speech-to-math}}.

\end{abstract}

\keywords{Automatic Speech Recognition, Natural Language Processing, Accessibility, \LaTeX\ Equations}

\maketitle

\section{Introduction}

\LaTeX\ is a widely used typesetting system in academia, especially in fields like mathematics, physics, and computer science, due to its ability to produce well-structured documents with precise formatting and complex mathematical notations. However, its intricate syntax often requires significant time and effort to master, posing a barrier to entry for many users. Unlike standard word processors, \LaTeX\ uses a markup language to define document structure and formatting, which can be daunting for beginners and inaccessible to those with disabilities. Visually impaired users, for example, may struggle to read and interpret \LaTeX\ code using screen readers or other assistive technologies. Similarly, users with motor impairments may find it challenging to input \LaTeX\ commands accurately, particularly when dealing with complex mathematical equations. These accessibility barriers limit the inclusivity of \LaTeX, as well as its adoption among a diverse range of users. Additionally, members of the visually impaired community worldwide have expressed that accessibility challenges present a significant obstacle to pursuing higher education and engaging in research and academia. In Greece, specifically, visually impaired individuals also encounter numerous challenges within educational institutions \citep{visually-impared-accessibility-greece}.

In light of these challenges, we focus our work on developing alternative approaches to interact with LaTeX that are more intuitive, accessible, and user-friendly. In this paper, we open source an end-to-end speech-to-text system specifically designed to generate \LaTeX\ equations (in source code format) based solely on audio input. In this way, the users are now able to verbally dictate mathematical expressions and equations, based on which our proposed system swiftly generates the respective \LaTeX\ equation. Our goal is to democratize access to \LaTeX\ and enhance the efficiency of mathematical communication for visually impaired people in Greece.

\section{Related Work}
\label{sec:background}

In the past, various approaches have been proposed to address the problem of transforming speech to structured mathematical equations, e.g. in \textit{MathML} or \LaTeX\ form. Initial approaches relied on rule-based conditions to generate mathematical equations from speech transcriptions \citep{mathifier, talkmaths}. It is notable that the ASR modules used in the aforementioned systems were often based on Hidden Markov Models (HMM), whose capabilities are limited compared to the current state of the art ASR systems.

An early approach, \textit{TalkMaths}, used Dragon Naturally Speaking (DNS) as the ASR system to transcribe speech. The transcriptions were mapped to mathematical notation using a customized Context-Free Grammar (CFG), which generated a parse tree that was then converted into the desired markup format. The authors also conducted statistical analyses on the frequency of unigrams, bigrams, and trigrams in the speech samples to develop a domain-specific Statistical Language Model (SLM). Additionally, they examined the impact of speech timing and prosody, aiming to establish a set of rules for reading mathematical expressions aloud.

Hanakovic et al. \citep{hanakovic-2006} created a system that categorizes serialized speech input into discrete mathematical categories, such as \textit{function}, \textit{symbol}, and \textit{fraction}. This approach converts speech into labeled text elements, for example, transforming “cosine of x plus three” into “function: cosine; parameter: x; symbol: plus; parameter: 3”. These labeled inputs are used to build an expression tree, which is then converted into concise mathematical notation, including \textit{MathML}. The system also includes \textit{navigation} and \textit{edit} modes, allowing users to select the category of the speech input or correct output errors directly. Moreover, Batlouni et al. \citep{mathifier} developed \textit{Mathifier}, a system designed to provide real-time speech recognition to help users retain long or complex equations. The system dynamically rendered the most probable output as speech was processed. They used \textit{Sphinx IV}\footnote{CMU Sphinx: https://cmusphinx.github.io}, developed by CMU, as the speech recognition module. To ensure mathematically sound outputs, they implemented a Grammar Language Model, which restricted the system’s output to domain-specific allowable sequences. This approach limited possible transitions during decoding to those that maintained mathematical coherence, such as preventing “minus” from being followed by “divided by”.


Recent advancements in deep learning have led to systems that can convert equations from images into structured textual form, like \LaTeX, benefiting visually or motor-impaired individuals. Wang et al. \citep{wang-img2tex} introduced an encoder-decoder architecture for this purpose. They used a Convolutional Neural Network (CNN) for encoding and a stacked bidirectional Long-Short Term Memory (LSTM) module with soft attention for decoding and token generation. The neural network undergoes two-step training: first, token-level training with Maximum-Likelihood Estimation (MLE), followed by sequence-level training using a reinforcement learning-based policy gradient algorithm, optimizing the model by considering the entire sequence of tokens.


\section{Dataset}
\label{sec:dataset}

Given the scarcity of publicly available domain-specific datasets and the limited knowledge of open-source ASR systems regarding both the Greek language and math-related speech nuances, we opted to develop our own task-specific dataset (henceforth denoted as \textit{Gr2Tex}). It consists of $500$ pairs of equations in natural text alongside their corresponding mathematical notation in \LaTeX\ form.

\smallskip \noindent \textbf{Dataset Collection Process}: The equations were selected from a variety of sources, including Greek mathematics textbooks, online educational platforms, and custom-generated examples crafted by computer science professionals. Our selection aimed to comprehensively cover mathematical concepts, from basic arithmetic to advanced algebra. For audio recordings, we engaged 10 native Greek speakers, both male and female, to read the equations aloud, ensuring minimal background noise and clear articulation of mathematical terms.

\smallskip \noindent \textbf{Data Preprocessing}: Prior to finalizing the dataset, we performed some preprocessing steps. The audio recordings were processed to restrict the background noise and normalized to a consistent volume level. The textual data was reviewed for consistency in notation, particularly ensuring uniform representation of symbols and expressions in LaTeX. We also standardized the text format, such as using consistent punctuation and spacing, to facilitate uniform processing during the model training phase. We split the dataset in train, validation and test set following a 70\%-15\%-15\% split.



\section{System Architecture}
\label{sec:architecture}

\begin{figure}[th!]
    \centering
    \includegraphics[width=\linewidth, trim={0.5cm 1.7cm 0.0cm 0.6cm}, clip]{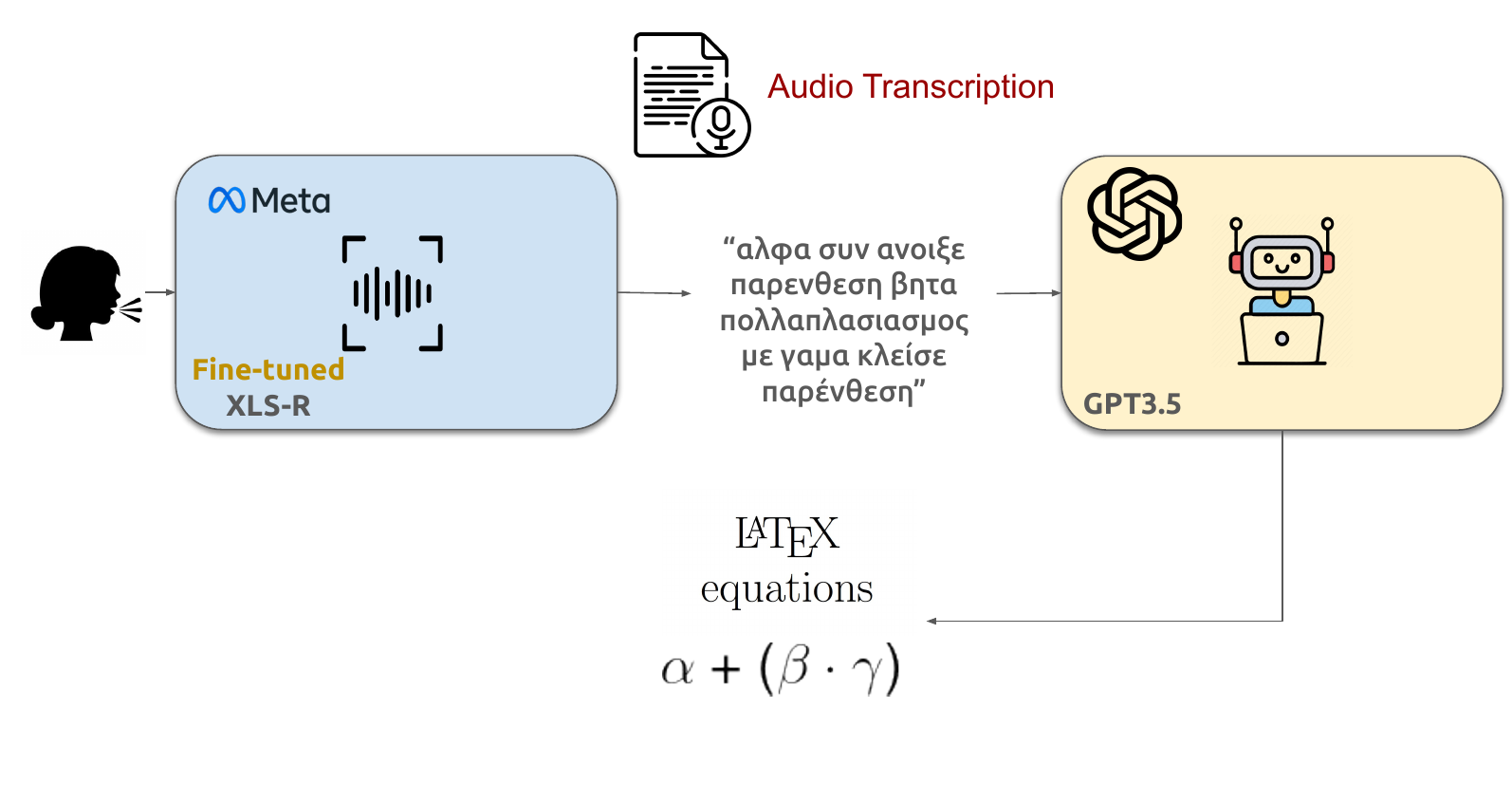}
\caption{\small An overview of the proposed system's architecture.}
\Description{An overview of the proposed system's architecture.}{}
\label{fig:1} 
\end{figure}

In this section, we provide further information about the architecture of our proposed system. Illustrated in Figure \ref{fig:1}, it consists of three primary components; a speech recognition module, a retrieval mechanism, as well as a text generation model. The speech recognition module transcribes speech input into free-form text, while the retrieval component searches a database to return the $k$ most similar equations in natural text from a held-out set of our dataset, along with their respective mathematical forms. Subsequently, the text generation model, in our case GPT-3.5-turbo, is prompted with the $k$ retrieved equation pairs, along with a brief instruction that guides the LLM's behaviour.

\subsection{Speech Recognition Component}
Our speech recognition component consisted of an instance of the XLS-R model developed by Meta AI \citep{xls-r}. While it is trained on thousands of hours of audio in multiple languages (including Greek), its performance on the Greek language was insufficient for integration into our proposed end-to-end pipeline. Thus, we underwent a fine-tuning process in order to initially refine the model's ability in successfully transcribing audio in Greek. Regarding the initial fine-tuning process, we collected approximately 26 hours of publicly available Greek speech audio, covering diverse topics such as political debates, sportcasting events and stand-up comedy shows. These recordings were transcribed and segmented into smaller chunks of approximately $3$ seconds each. Additionally, we utilized a subset of the Common Voice \citep{common-voice} dataset that contained Greek audio. Finally, we also fine-tuned the model using our custom domain- and language-specific dataset (see Section \ref{sec:dataset}) in order to enhance its ability to accurately transcribe math-related audio inputs.

\subsection{Equation Generation Component}

Given the rapid advancements in the NLP domain, we chose to use a Large Language Model (LLM) for the equation generation process. However, a notable challenge emerges due to the limited availability of LLMs pretrained on a substantial Greek language corpus, with even fewer trained on math-related data. Our decision to develop a speech-to-LaTeX tool for the Greek language meant that models specifically tailored for code generation, such as StarCoder \citep{starcoder} could not be employed, as their capabilities in Greek are limited. While such models would excel due to the code-like structure of \LaTeX\ equations, they lack sufficient training data for Greek language tasks. Thus, we decided to employ OpenAI's GPT3.5-turbo. 

A common strategy when leveraging LLMs is to enhance their capabilities through in-context learning (ICL). In the NLP domain, ICL is a paradigm where an LLM is prompted with a set of instructions alongside a few task-specific demonstrative examples. This allows the model to adapt its output based solely on the provided examples, without the need to update its parameters \citep{icl-survey}. Given the lack of a task-specific dataset for Greek, aside from our own \textit{Gr2Tex} with only 500 examples, ICL was essential for guiding the LLM’s responses. We selected demonstrative examples by retrieving the $k$ most semantically similar samples to the query, with $k$ being a hyperparameter optimized during a tuning process. Balancing performance with the cost of using a closed-source LLM like GPT-3.5, we experimented with values of $k$ from two to six, as detailed in Section \ref{sec:results}.

\subsection{Retrieval Mechanism}

Our retrieval mechanism operates on the \textit{Gr2TeX} dataset (see Section \ref{sec:dataset}), composed of a diverse range of mathematical equations represented in natural language and their corresponding LaTeX forms. We employ the $k$-NN algorithm in order to identify the $k$ most similar examples to the current query. We experimented with various similarity and distance measures, such as the cosine and euclidean similarity, as well as the Manhattan distance. The results of our exploration, detailing which similarity or distance function yielded better performance, are presented in Section \ref{sec:results}.

\section{Experimental Setup}

\subsection{Evaluation Metrics}
\label{sec:evaluation}

We evaluated our system’s performance by measuring the Levenshtein distance between the generated equations and ground truth, following preprocessing steps to standardize the equations. This preprocessing involved removing LaTeX-specific formatting and commands, such as dollar signs and equation delimiters, and standardizing mathematical symbols and variables, including replacing Greek letters with Latin equivalents and removing unnecessary punctuation. After preprocessing, the function calculates the Levenshtein distance, which quantifies the minimum number of single-character edits required to transform one string into another. We assessed the effectiveness of this evaluation method, which we denote as \textit{EL}, by comparing its results to a set of annotated assessments conducted by a group of five individuals. Each pair of generated and ground truth equations was categorized as either ``Not Match", ``Almost Match", or ``Match", denoted respectively as $-1$, $0$, and $1$. In addition to this evaluation method, we report typical Natural Language Generation (NLG) metrics such as BLEU \citep{papineni-etal-2002-bleu} and chrF \citep{popovic-2015-chrf}.

\subsection{Experimental Results}
\label{sec:results}

In this section, we present the outcomes of our experiments and discuss our observations. Table \ref{fig:results} showcases the performance of our proposed system across different values of $k$ and various similarity or distance functions. The first row of the table corresponds to our baseline approach, which did not utilize any in-context learning (ICL) functionality. In this setup, the generative model was prompted solely to produce the \LaTeX\ equation for the provided transcribed text sequence. The third column indicates the selected prompt (see Table 2). The fourth and fifth columns of the table indicate the percentage of equations with an EL distance lower than $0.1$ or $0.4$ from their respective ground truths. The final two columns present the BLEU \citep{papineni-etal-2002-bleu} and chrF \citep{popovic-2015-chrf} scores respectively.

\begin{table}[h]
    \centering
    \begin{tabular}{|p{0.5cm}|p{7.2cm}|}
     \hline
     \multicolumn{2}{|c|}{\textbf{Instruction Prompts}} \\
     \hline
     \centering $p1$ & {\small ``You are a LaTeX equation generator. You are provided with an equation described in natural text and you are asked to generate the respective LaTeX equation.''}\\
     \hline
     \centering $p2$ & {\small ``You are a LaTeX equation generator. You are provided with an equation described in natural text and you are asked to generate the respective LaTeX equation. Follow the examples and generate the LaTeX equation for the last query.''}\\
     \hline
     \centering $p3$ & {\small ``\textgreek{Είσαι ένας βοηθός προγραμματιστή. Σου παρέχεται μία εξίσωση σε φυσική γλώσσα και σου ζητείται να παράξεις την αντίστοιχη εξίσωση σε κώδικα} LaTeX. \textgreek{Συμπλήρωσε την εξίσωση σε κώδικα} LaTeX \textgreek{για το τελευταίο αίτημα.}''}\\
     \hline
    \end{tabular}
    \caption{\label{table:instructios} \small The instruction prompts used throughout our experiments.}
\end{table}

\begin{table}[h!]
    \centering
    \begin{tabular}{|p{0.4cm}|p{1.15cm}|p{0.8cm}||p{0.98cm}|p{0.98cm}|p{0.85cm}|p{0.85cm}|}
        \hline
         \multicolumn{7}{|c|}{\small\textbf{Results on Gr2TeX - GPT3.5}} \\
        \hline
        \cline{1-7}
        \centering\small \textit{k} &\centering\small \textit{Sim/Dist} &\centering\small \textit{Prompt} & \centering\small \textit{EL $< 0.1$} & \centering\small \textit{EL $> 0.4$} & \centering\small \textit{BLEU} & \small \hspace{0.1cm} \textit{chrF} \\
        \hline
        \cline{1-7}
        \small \centering - & \small \centering - & \small \centering - & \small \centering 27.45 & \small \centering 24.84 & \small \centering 39.54 &  \hspace{0.1cm}\small 60.58\\
        \hline
        \small \centering 3 & \small \centering Cosine & \small \centering $p1$ & \small \centering 32.06 & \small \centering 29.85 & \small \centering 39.77 &  \hspace{0.1cm}\small 61.86\\
        \hline
        \small \centering 3 & \small \centering Cosine & \small \centering $p3$ & \small \centering 34.66 & \small \centering 23.04 & \small \centering 42.37 &  \hspace{0.1cm}\small 63.92 \\
        \hline
        \small \centering 4 & \small \centering Euclidean & \small \centering $p2$ & \small \centering 34.55 & \small \centering 24.27 & \small \centering 44.88 &  \hspace{0.1cm}\small 63.26\\
        \hline
        \small \centering 5 & \small \centering Manhattan & \small \centering $p1$ & \small \centering 36.03 & \small \centering 21.01 & \small \centering 47.95 & \hspace{0.1cm}\small 65.77 \\
        \hline
        \small \centering 5 & \small \centering Manhattan & \small \centering $p2$ & \small \centering 36.15 & \small \centering 20.84 & \small \centering \textbf{53.42} & \hspace{0.1cm}\small 66.03 \\
        \hline
        \small \centering 5 & \small \centering Cosine & \small \centering $p2$ & \small \centering \textbf{37.67} & \small \centering \textbf{17.03} & \small \centering 52.33 &  \hspace{0.1cm}\small 66.17 \\
        \hline
        \small \centering 5 & \small \centering Cosine & \small \centering $p3$ & \small \centering 35.98 & \small \centering 21.04 & \small \centering 48.21 &  \hspace{0.1cm}\small 64.38 \\
        \hline
        \small \centering 6 & \small \centering Manhattan & \small \centering $p2$ & \small \centering 34.86 & \small \centering 21.04 & \small \centering 46.79 & \hspace{0.1cm}\small 63.69 \\
        \hline
        \small \centering 6 & \small \centering Cosine & \small \centering $p2$ & \small \centering 37.59 & \small \centering 17.51 & \small \centering 52.51 & \hspace{0.1cm}\small \textbf{66.24} \\
        \hline
    \end{tabular}
    \\
    \caption{\small The results of our experiments on \textit{Gr2TeX} using GPT3.5.}
    \label{fig:results}
\end{table}

We observed a significant impact of the number of demonstrative examples on the performance of the employed LLM. Specifically, we noticed a discernible difference in performance between scenarios with $k=3$ and $k=5$. However, increasing the number of examples beyond a certain point (specifically $k$ = $6$) did not enhance performance. Due to the costs associated with using closed-source LLMs like GPT-3.5, we did not test with more examples, but this is planned for future research to further validate our findings. 

Furthermore, the choice of instruction prompt significantly influences the system's performance. In our experiments, we tested three different instructions, outlined in Table 2. Notably, instruction $p2$ demonstrated superior effectiveness compared to instruction $p3$, which is almost the same as $p2$ but in Greek. Specifically, $p2$ improved the $\textit{EL} < 0.1$ metric by up to 1.5\% and reduced the $\textit{EL} > 0.4$ metric by up to 3.8\%, which is the desirable outcome. Similarly, instruction $p1$ generally underperformed compared to the other two prompts. This underscores the importance of exploring different prompting strategies to optimize system performance. In addition, the cosine similarity measure seems to lead to slightly better overall performance based on the results of Table \ref{fig:results}.

\section{Web Application}
\label{sec:webapp}

\begin{figure}[h!]
  \centering
  \begin{tabular}[b]{c}
    \includegraphics[height=6.0cm, width=0.9\linewidth]{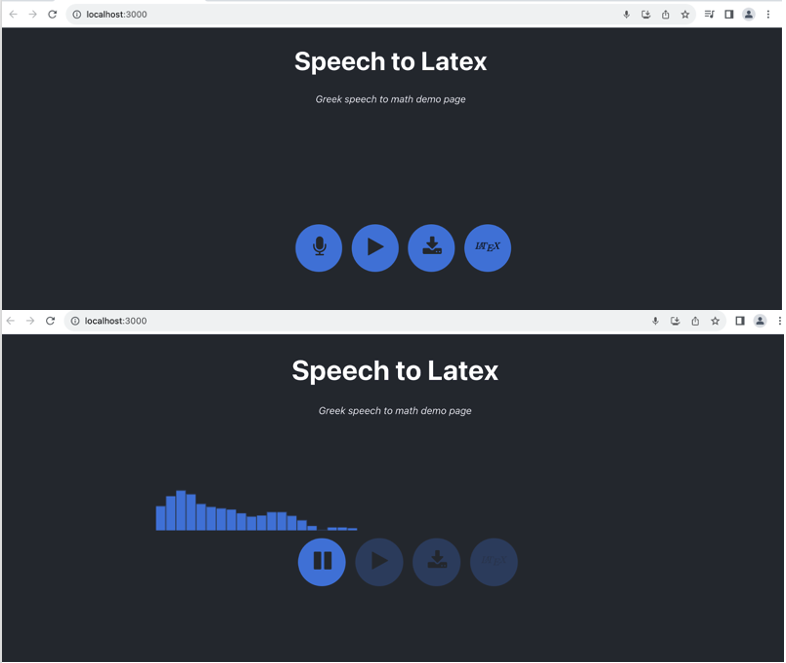} \\
     \includegraphics[width=0.9\linewidth]{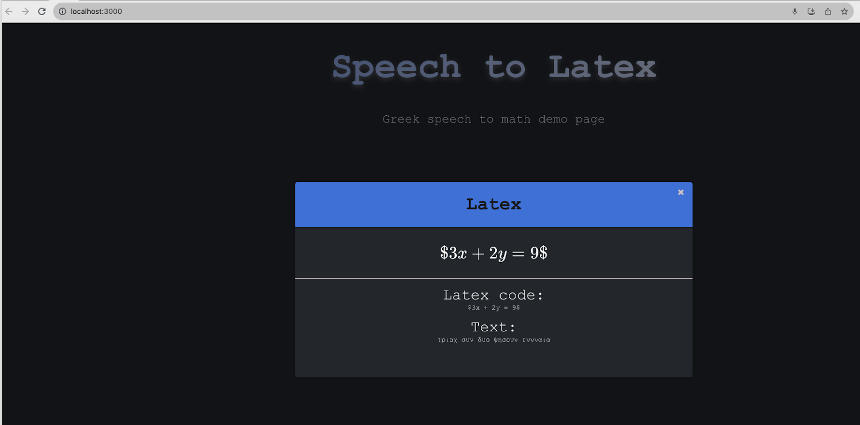} \\
  \end{tabular}
  \caption{\small The UI when entering the application and during recording. The modal window appears when clicking on the \LaTeX\ button.}
  \Description{The UI when entering the application and during recording. The modal window appears when clicking on the \LaTeX\ button.}{}
  \label{fig:2}
\end{figure}

To address accessibility barriers in formulating mathematical expressions, we integrated $Gr2tex$ into a web application, with the backend developed in FASTAPI and the frontend in React. The interface includes four main buttons: one for recording the mathematical expression, another for playing back the recorded phrase, a third for downloading the audio file, and a fourth for converting speech to \LaTeX\ (Figure \ref{fig:2}). Clicking the \LaTeX\ button transcribes the audio and sends the text to the model that generates the corresponding LaTeX expression. This expression is displayed in a modal window, giving users immediate access to the mathematical representation of their spoken input (Figure \ref{fig:2}). The web application is accessible by following the instructions found in our open-source \href{https://github.com/magcil/greek-speech-to-math}{\textcolor{blue}{repository}}.

\section{Conclusions \& Future Work}
\label{sec:conclusions}

In this work, we introduced an end-to-end speech-to-text system designed to generate \LaTeX\ equations. The system comprises an ASR module, a text generation component, and an assistive retrieval mechanism. The ASR module is a fine-tuned version of the XLS-R model \citep{xls-r}, while the text generation component utilizes GPT3.5. We leverage in-context learning to dynamically enhance the model’s performance without further training, using a set of demonstrative examples retrieved for each query from a held-out set. This system represents a significant advancement in translating spoken mathematical expressions into \LaTeX\ format, improving accessibility and efficiency in scientific communication. In future work, our objective is to explore more advanced ASR systems and newer LLMs with enhanced capabilities in Greek language processing. Additionally, we plan to investigate more sophisticated retrieval techniques and experiment with elaborate prompting strategies, such as Chain-of-Thought (CoT) prompting.

\bibliographystyle{ACM-Reference-Format}

\footnotesize
 \bibliography{sample-base}


\begin{thebibliography}{11}


\ifx \showCODEN    \undefined \def \showCODEN     #1{\unskip}     \fi
\ifx \showDOI      \undefined \def \showDOI       #1{#1}\fi
\ifx \showISBNx    \undefined \def \showISBNx     #1{\unskip}     \fi
\ifx \showISBNxiii \undefined \def \showISBNxiii  #1{\unskip}     \fi
\ifx \showISSN     \undefined \def \showISSN      #1{\unskip}     \fi
\ifx \showLCCN     \undefined \def \showLCCN      #1{\unskip}     \fi
\ifx \shownote     \undefined \def \shownote      #1{#1}          \fi
\ifx \showarticletitle \undefined \def \showarticletitle #1{#1}   \fi
\ifx \showURL      \undefined \def \showURL       {\relax}        \fi
\providecommand\bibfield[2]{#2}
\providecommand\bibinfo[2]{#2}
\providecommand\natexlab[1]{#1}
\providecommand\showeprint[2][]{arXiv:#2}

\bibitem[Ardila~et al.(2020)]%
        {common-voice}
\bibfield{author}{\bibinfo{person}{R. Ardila~et al.}} \bibinfo{year}{2020}\natexlab{}.
\newblock \showarticletitle{Common Voice: A Massively-Multilingual Speech Corpus}. In \bibinfo{booktitle}{\emph{Proceedings of the Twelfth Language Resources and Evaluation Conference}}. \bibinfo{address}{Marseille, France}.
\newblock


\bibitem[Babu et~al\mbox{.}(2021)]%
        {xls-r}
\bibfield{author}{\bibinfo{person}{A. Babu}, \bibinfo{person}{C. Wang}, \bibinfo{person}{A. Tjandra}, \bibinfo{person}{K. Lakhotia}, \bibinfo{person}{Q. Xu}, \bibinfo{person}{N. Goyal}, \bibinfo{person}{K. Singh}, \bibinfo{person}{P. von Platen}, \bibinfo{person}{Y. Saraf}, \bibinfo{person}{J.~M. Pino}, \bibinfo{person}{A. Baevski}, \bibinfo{person}{A. Conneau}, {and} \bibinfo{person}{M. Auli}.} \bibinfo{year}{2021}\natexlab{}.
\newblock \showarticletitle{XLS-R: Self-supervised Cross-lingual Speech Representation Learning at Scale}. In \bibinfo{booktitle}{\emph{Interspeech}}.
\newblock


\bibitem[Batlouni et~al\mbox{.}(2011)]%
        {mathifier}
\bibfield{author}{\bibinfo{person}{S. Batlouni}, \bibinfo{person}{H. Karaki}, \bibinfo{person}{F. Zaraket}, {and} \bibinfo{person}{F. Karameh}.} \bibinfo{year}{2011}\natexlab{}.
\newblock \showarticletitle{Mathifier — Speech recognition of math equations}. In \bibinfo{booktitle}{\emph{2011 18th IEEE International Conference on Electronics, Circuits, and Systems}}. \bibinfo{pages}{301--304}.
\newblock


\bibitem[Dong et~al\mbox{.}(2023)]%
        {icl-survey}
\bibfield{author}{\bibinfo{person}{Q. Dong}, \bibinfo{person}{L. Li}, \bibinfo{person}{D. Dai}, \bibinfo{person}{C. Zheng}, \bibinfo{person}{Z. Wu}, \bibinfo{person}{B. Chang}, \bibinfo{person}{X. Sun}, \bibinfo{person}{J. Xu}, \bibinfo{person}{L. Li}, {and} \bibinfo{person}{Z. Sui}.} \bibinfo{year}{2023}\natexlab{}.
\newblock \bibinfo{title}{A Survey on In-context Learning}.
\newblock
\newblock
\showeprint[arxiv]{2301.00234}


\bibitem[et~al.(2023)]%
        {starcoder}
\bibfield{author}{\bibinfo{person}{R.~Li et al.}} \bibinfo{year}{2023}\natexlab{}.
\newblock \showarticletitle{StarCoder: may the source be with you!}
\newblock \bibinfo{journal}{\emph{Transactions on Machine Learning Research}} (\bibinfo{year}{2023}).
\newblock


\bibitem[Hanakovi{\v{c}} and Nagy(2006)]%
        {hanakovic-2006}
\bibfield{author}{\bibinfo{person}{T. Hanakovi{\v{c}}} {and} \bibinfo{person}{M. Nagy}.} \bibinfo{year}{2006}\natexlab{}.
\newblock \showarticletitle{Speech Recognition Helps Visually Impaired People Writing Mathematical Formulas}. In \bibinfo{booktitle}{\emph{Computers Helping People with Special Needs}}. \bibinfo{publisher}{Springer Berlin Heidelberg}, \bibinfo{address}{Berlin, Heidelberg}, \bibinfo{pages}{1231--1234}.
\newblock


\bibitem[Katsoulis(2008)]%
        {visually-impared-accessibility-greece}
\bibfield{author}{\bibinfo{person}{P. Katsoulis}.} \bibinfo{year}{2008}\natexlab{}.
\newblock \showarticletitle{The current educational situation for students with visual impairment in Greece: Trends and prospects}.
\newblock \bibinfo{journal}{\emph{“Recent Approaches \& Future Challenges: Programs and Projects regarding the VI \& MDVI”}} (\bibinfo{year}{2008}).
\newblock


\bibitem[Papineni et~al\mbox{.}(2002)]%
        {papineni-etal-2002-bleu}
\bibfield{author}{\bibinfo{person}{K. Papineni}, \bibinfo{person}{S. Roukos}, \bibinfo{person}{T. Ward}, {and} \bibinfo{person}{W. Zhu}.} \bibinfo{year}{2002}\natexlab{}.
\newblock \showarticletitle{{B}leu: a Method for Automatic Evaluation of Machine Translation}. In \bibinfo{booktitle}{\emph{Proceedings of the 40th Annual Meeting of the Association for Computational Linguistics}}. \bibinfo{publisher}{Association for Computational Linguistics}, \bibinfo{address}{Philadelphia, Pennsylvania, USA}.
\newblock


\bibitem[Popovi{\'c}(2015)]%
        {popovic-2015-chrf}
\bibfield{author}{\bibinfo{person}{M. Popovi{\'c}}.} \bibinfo{year}{2015}\natexlab{}.
\newblock \showarticletitle{chr{F}: character n-gram {F}-score for automatic {MT} evaluation}. In \bibinfo{booktitle}{\emph{Proceedings of the Tenth Workshop on Statistical Machine Translation}}. \bibinfo{publisher}{Association for Computational Linguistics}, \bibinfo{address}{Lisbon, Portugal}.
\newblock


\bibitem[Wang and Liu(2021)]%
        {wang-img2tex}
\bibfield{author}{\bibinfo{person}{Z. Wang} {and} \bibinfo{person}{J. Liu}.} \bibinfo{year}{2021}\natexlab{}.
\newblock \showarticletitle{Translating math formula images to LaTeX sequences using deep neural networks with sequence-level training}.
\newblock \bibinfo{journal}{\emph{International Journal on Document Analysis and Recognition (IJDAR)}}  \bibinfo{volume}{24} (\bibinfo{date}{06} \bibinfo{year}{2021}), \bibinfo{pages}{1--13}.
\newblock


\bibitem[Wigmore et~al\mbox{.}(2009)]%
        {talkmaths}
\bibfield{author}{\bibinfo{person}{A. Wigmore}, \bibinfo{person}{G. Hunter}, \bibinfo{person}{E. Pfluegel}, \bibinfo{person}{J. Denholm-Price}, {and} \bibinfo{person}{V. Binelli}.} \bibinfo{year}{2009}\natexlab{}.
\newblock \showarticletitle{Using Automatic Speech Recognition to Dictate Mathematical Expressions: The Development of the 'TalkMaths' Application at Kingston University}.
\newblock  (\bibinfo{date}{01} \bibinfo{year}{2009}).
\newblock


\end{thebibliography}






\end{document}